\documentclass[11pt]{article}

\usepackage[final]{acl}

\usepackage{times}
\usepackage{latexsym}

\usepackage[T1]{fontenc}

\usepackage[utf8]{inputenc}

\usepackage{microtype}

\usepackage{inconsolata}

\usepackage{hyperref}
\usepackage{url}
\usepackage{graphicx}
\graphicspath{ {./figures/} }
\usepackage{subcaption}
\usepackage{tikz}
\usepackage{pgffor}
\usetikzlibrary{matrix, positioning, calc, decorations.pathreplacing, patterns}
\usepackage{amsmath}
\usepackage{amssymb}
\usepackage{booktabs}
\usepackage{algorithm}
\usepackage{algpseudocode}
\usepackage{xpatch}
\usepackage{enumitem}
\usepackage{multirow}
\usepackage{array}
\usepackage{bm}
\usepackage{tablefootnote}
\usepackage{tcolorbox}
\tcbuselibrary{skins, breakable, theorems}

\title{ChunkAttention: Efficient Self-Attention with Prefix-Aware KV Cache and Two-Phase Partition}

\author{Lu Ye \quad Ze Tao \quad Yong Huang \quad Yang Li \\
 Microsoft\\
 \small \texttt{\{luye,zetao,yohuan,yali2\}@microsoft.com}}

\begin{document}
\maketitle

\begin{abstract}
Self-attention is an essential component of large language models (LLM) but a significant source of inference latency for long sequences. In multi-tenant LLM serving scenarios, the compute and memory operation cost of self-attention can be optimized by using the probability that multiple LLM requests have shared system prompts in prefixes. In this paper, we introduce ChunkAttention, a prefix-aware self-attention module that can detect matching prompt prefixes across multiple requests and share their key/value tensors in memory at runtime to improve the memory utilization of KV cache. This is achieved by breaking monolithic key/value tensors into smaller chunks and structuring them into the auxiliary prefix tree. Consequently, on top of the prefix-tree based KV cache, we design an efficient self-attention kernel, where a two-phase partition algorithm is implemented to improve the data locality during self-attention computation in the presence of shared system prompts. Experiments show that ChunkAttention can speed up the self-attention kernel by 3.2-4.8$\times$ compared to the state-of-the-art implementation, with the length of the system prompt ranging from 1024 to 4096. \footnote{Code is publicly available at \url{https://github.com/microsoft/chunk-attention}}
\end{abstract}

\section{Introduction}

Over the last few years, Large Language Models (LLM) have developed various capabilities, from in-context learning \citep{iclsurvey} to chain-of-thought reasoning \citep{cotsurvey,wei2022chain}, and achieved remarkable success in a wide range of natural language processing related tasks \citep{chang2023survey}. Representive models are the GPT \citep{gpt1,gpt2,gpt3,gpt4}, LLaMA \citep{llama}, PaLM \citep{palm2} and Gemini \citep{geminiteam2023gemini} series. Following the success of ChatGPT and GPT store, LLM-based applications start to surge, and the demand to optimize LLM's inference cost has been a new area of research interest \citep{kim2023stack,flexgen,10046087}.

The self-attention module, as one of the critical components in LLMs, has poor performance during inference (Table~\ref{tab:llm_cost_breakdown}) since it performs intensive memory operations on key/value tensors of context tokens (KV cache) and is memory-bound \citep{roofline,jin2023s}. The memory complexity grows linearly with context length. As the demand for more context tokens has been a trend (32K for GPT-4), the performance gets worse \citep{gpt4}. KV cache additionally restricts the batch size and system throughput. For instance, using FP16, the KV cache for each token in GPT-3(175B) requires 4.5MB of memory. The memory of an inference server with 8*A100 (80G) can only hold 70000 tokens or 35 sequences of 2K context tokens.

On the other hand, the shared system prompt in LLM-based applications leads to redundancy in KV cache \citep{anthropic:online}. Typically, LLMs are pre-trained and deployed in a multi-tenant architecture for multiple applications to share. Due to the in-context learning abilities of LLMs, using the system prompt to guide LLMs with instructions and few-shot examples is a common practice in designing LLM-based applications \citep{white2023prompt,zhou2023large}. The system prompt is shared between multiple requests and can be very long. This can be observed in various LLM-based applications, from online chatbots to offline experiments (\S~\ref{sec:sys_prompt}). For instance, \citet{leakedsystemprompts} shows that system prompts of various LLM-based applications have more than 1K tokens. For a ChatGPT-like online chatbot, the system prompt can be as long as 1766 tokens with only 6 plugins activated, and all requests sent to the chatbot share one single system prompt (Appendix \ref{app:sys_prompt}). In the Chameleon \citep{lu2023chameleon} work, 4 system prompts are shared by 4241 queries to run the ScienceQA benchmark \citep{lu2022learn}, and 7 system prompts are shared by 7685 queries to run the TabMWP benchmark \citep{lu2022dynamic}.

An important question is whether we can leverage the sharing characteristic of system prompts to make the self-attention module faster and more memory efficient. To our knowledge, the only related work is a proposal by \citet{vllm}, in which the service provider reserves memory for key/value tensors of a set of predefined system prompts from application developers. The proposal has limitations: i) predefined system prompts are static and inflexible in frequent refreshes for large-scale deployments since both application developers and the service provider are involved in the operation loop; ii) there is memory waste in case of long system prompts and low hit rate; iii) no work has been done to optimize the self-attention kernel in the presence of shared system prompts.

To fill the gap, we propose ChunkAttention, a novel self-attention module featuring the prefix-aware KV cache (PAKV) and two-phase partition (TPP). KV cache in ChunkAttention is a prefix tree built with chunked context tokens and key/value tensors. Thus, the KV cache is prefix-aware and can dynamically detect and remove redundancy at runtime without human involvement. The KV cache only stores key/value tensors of sequences currently in decoding and has zero memory waste. In addition, the prefix-tree structure provides context for ChunkAttention to redesign a highly-optimized self-attention kernel with two-phase partition: chunk-first phase and sequence-first phase. Query tensors from sequences with matching prompt prefixes are batched together to perform attention with key/value tensors.

The main contributions of this paper are as follows: i) we reveal that system prompts can be long (\S \ref{sec:sys_prompt}), providing opportunities for optimizing self-attention; ii) we propose to use prefix tree to implement KV cache, which is out-of-the-box, scalable and robust in terms of redundancy removal; iii) we implement a two-phase partition algorithm to speed up self-attention kernel on prefix-aware KV cache; iv) we prove the feasibility and empirically quantify the gain self-attention can achieve from shared system prompts under various system configurations. Our experiments show that ChunkAttention can be significantly faster as the length of shared system prompts grows and has no performance degradation without shared system prompts, compared to existing highly optimized implementations.

\begin{table}[t]
    \centering
    \resizebox{\columnwidth}{!}{
    \begin{tabular}{ c| c | ccc}
    \toprule
    Batch Size & Roofline & QKV Projection & Self Attention  & MLP \\
                    
    \midrule
    \multirow{3}{*}{1} & FLOPs($\times 10^6$) & 100.66 & 33.57 & 270.53 \\
     & MOPs($\times 10^6$) & 100.70 & 33.85 & 270.62 \\
     & Arithmetic Intensity & 1.00 & \textbf{0.99} & 1.00 \\
     & Latency(\textmu s) & 88.44 & \textbf{17.82} & 160.77 \\
    \midrule
    \multirow{3}{*}{32} & FLOPs($\times 10^6$) & 3221.23 & 1074.27 & 8657.04 \\
     & MOPs($\times 10^6$) & 101.71 & 1083.18 & 273.43 \\
     & Arithmetic Intensity & 31.67 & \textbf{0.99} & 31.66 \\
     & Latency(\textmu s) & 90.02 & \textbf{687.74} & 209.82 \\
    \midrule
    \multirow{3}{*}{64} & FLOPs($\times 10^6$) & 6442.45 & 2148.53 & 17314.09 \\
     & MOPs($\times 10^6$) & 102.76 & 2166.36 & 276.33 \\
     & Arithmetic Intensity & 62.69 & \textbf{0.99} & 62.66 \\
     & Latency(\textmu s) & 98.04 & \textbf{1358.40} & 217.79 \\
    \bottomrule
    \end{tabular}
    }
    \caption{Complexity analysis of key modules in each decoder layer when decoding one single token. Llama2 7B, 2048 context tokens, FP16, A100 (80G). The self-attention module has low arithmetic intensity \citep{roofline} and high latency. FLOPs: floating point operations. MOPs: memory operations or memory bytes accessed. Arithmetic Intensity: FLOPs/MOPs.}
    \label{tab:llm_cost_breakdown}
\end{table}

\section{Preliminaries}

\subsection{Shared System Prompt} \label{sec:sys_prompt}

One paradigm in designing LLM-based applications has been the introduction of system prompt \citep{anthropic:online}. It provides instructions, few-shot examples \citep{iclsurvey}, and external knowledge as context for LLMs to generate better results. The final prompt to LLMs is a concatenation of system prompt and task-specific input. The system prompt is shared between multiple requests and can be very long. This can be observed in various LLM-based applications, from online chatbots to offline experiments.

Toolformer or using external tools becomes an essential skill for LLMs to get up-to-date information or perform precise math calculations \citep{schick2023toolformer,li2023apibank}. It is implemented by plugins in ChatGPT-like online chatbot applications \citep{ChatGPTPlugins:online}. Equivalent capability is provided by GPT series models through function calling \citep{ChatGPTFuncCallDoc:online}. Under the hood, available function specifications are silently injected into the system prompt \cite{ChatGPTFuncCallBlog:online}. Experiments indicate that with 6 plugins activated, the token length of the shared system prompt can reach up to 1766 (Appendix \ref{app:sys_prompt}).

Another source of shared system prompts is the offline research-focused experiments conducted on LLMs. In these scenarios, researchers frequently create a large number of templated requests with identical instructions, examples, or external knowledge and issue them to LLMs quickly. Example work includes: i) Chameleon \citep{lu2023chameleon} reuses policy planning and tool invocation prompts for compositional reasoning on the ScienceQA and TabMWP datasets; ii) CREATOR \cite{qian2023creator} constructs a collection of questions from TabMWP and MATH datasets using a chain-of-thought (CoT) prompt template; iii) PDFTriage \cite{saadfalcon2023pdftriage} injects the PDF document metadata into prompt and runs multiple question-answering (QA) tasks over the document; iv) ToolQA \cite{zhuang2023toolqa} further releases a QA dateset and reuses the system prompt for evaluations of QA with LLMs. Table \ref{tab:shared_sys_prompt} shows statistics on shared token counts of system prompts.

\footnotetext[1]{\url{https://github.com/lupantech/chameleon-llm/blob/main/run_tabmwp/demos/prompt_policy.py}}
\footnotetext[2]{\url{https://github.com/qiancheng0/CREATOR/blob/main/MATH/prompt_lib/prompt_cot.md}}
\footnotetext[3]{\url{https://github.com/night-chen/ToolQA/blob/main/benchmark/chameleon/run_toolqa/demos/prompt_policy.py}}

\begin{table}
    \centering
    \resizebox{\columnwidth}{!}{
    \begin{tabular}{llrrrrrr}
    \toprule
    \multirow{2}{*}{System} & \multirow{2}{*}{Usage of Prompt} & \multicolumn{2}{c}{\#shared prompt tokens}  \\ \cmidrule(lr){3-4}
    & & avg & max \\
    \midrule
    {Chameleon} & {Tools definition and examples \footnotemark[1]} & {1324} & {2626} \\
    {CREATOR} & {CoT examples \footnotemark[2]} & {879} & {2492} \\        
    {PDFTriage} & {PDF document metadata} & {4257} & {N.A.} \\
    {ToolQA} & {Tools definition and examples \footnotemark[3]} & {1432} & {1432} \\
    \bottomrule
    \end{tabular}
    }
    \caption{Shared prompt tokens in system prompt, tokenized by OpenAI's tiktoken tokenizer library \cite{openaitiktok:online}.}
    \label{tab:shared_sys_prompt}
\end{table}

\subsection{LLM Inferencing}

The typical inference process of LLMs consists of two stages: prefilling and decoding \citep{flexgen}. After receiving a sequence $S=[t_1,...,t_{n_p}]$, the server starts to prefill. During prefilling, it feeds all $n_p$ prompt tokens $t_1,...,t_{n_p}$ into LLMs, computes the attention key/value tensors, and caches them to speed up subsequent computations. Then, the server performs decoding. Decoding is auto-regressive, and the input token to LLMs is the completion token(or output token) generated from the previous decoding iteration. The process continues until the end-of-sequence token or maximum completion tokens are generated.

When the server is decoding $b$ (batch size) sequences $S_1,...,S_b$ simultaneously, although they are in different iterations, the server can still perform batching at the granularity of iteration and predict the next tokens for all sequences together, rather than separately, which is known as iteration-based batching \citep{gao2018low,orca,ebatch}. Specifically, iteration-based batching concatenates last input tokens of multiple sequences (one token per sequence) $t^{(1)},...,t^{(b)}(t^{(i)} \in S_i)$ into a single input $\bm{T}$, and computes the QKV projection before self-attention, the output projection and multilayer perceptron after self-attention. The self-attention in the middle has no shared weights and needs to be computed independently for each sequence. During decoding, new sequences can join, and completed sequences can leave, significantly increasing the possibility of forming big batches. Iteration-based batching has been implemented by vLLM \citep{vllm} and the text-generation-inference server \citep{tgi}. The ChunkAttention in this paper assumes that iteration-based batching is enabled to form batches for its kernel to run efficiently.

\section{Our Approach}

\subsection{Prefix Aware KV Cache (PAKV)} \label{sec:PAKV}

\begin{figure*}[htb]
  \centering
  \resizebox{0.9\linewidth}{!} {
      \input{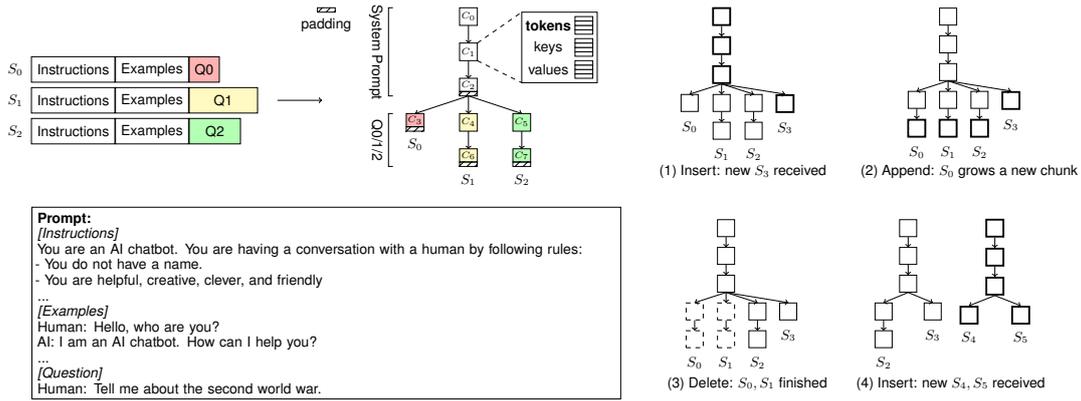}
  }
  \caption{KV cache in prefix tree. The instructions and examples in prompts of $S_0, S_1, S_2$ are common and sharable. Questions are different and not sharable. Some memory is unused due to alignment.}
  \label{fig:chunk_attn_kvcache}
\end{figure*}

Traditionally, KV cache is stored in dense tensors of size $b \times h \times n \times d$ where $b$ is the batch size, $h$ is the number of heads, $n$ is the sequence length, and $d$ is the head dimension size. 

When multiple sequences share common prefix tokens, key/value tensors are the same and thus can be shared in memory. For example, a particular LLM inference server receives sequence $S_i=[t_1,...,t_{n_s},t_{n_s+1},...,t_{n_p}]$ first, and then receives sequence $S_j=[t_1,...,t_{n_s},t_{n_s+1}',...,t_{n_p}']$. KV cache for $t_1,...,t_{n_s}$ can only have one physical copy in memory. 

Given the property, we argue that the KV cache should be made prefix-aware, which is to organize the KV cache of all sequences under decoding into a prefix tree. Precisely, we slice monolithic key/value tensors contiguous in memory along the sequence length dimension. Figure~\ref{fig:chunk_attn_kvcache} shows the structure of the KV cache stored in a prefix tree. Each node defines a chunk $C$ storing three essential elements: i) a segment of $c$ context tokens shared by sequences $S_i,..., S_j$ to enable prefix tree operations; ii) a slice of key tensor of size $b \times h \times c \times d$ for the $c$ tokens; iii) the corresponding slice of value tensor. Each path in the prefix tree defines a sequence. Multiple trees (a forest) may exist in the server simultaneously. For instance, application developers design different system prompts.

There are three possible scenarios during inference: i) new sequence joins, ii) completed sequence leaves, and iii) all sequences decode one token together. Each scenario can be translated into prefix tree operations. When a new sequence joins, the prefix tree is searched and updated to insert a new path. When a completed sequence leaves, the prefix tree is updated to delete its path. At each decoding iteration, we append new tokens into leaf chunks or grow a new chunk when the leaf chunk is full.

Given a fixed chunk size $c$, memory management is efficient. In ChunkAttention, the pool-based memory allocator is adopted by default \citep{HILL199249,poolallocator}. It keeps track of both a used and a free chunk list. When a new chunk is requested, the allocator returns a chunk from the free list or allocates fresh memory from the operating system (OS). Unused chunks are returned to the allocator once a sequence is completed, but the allocator does not release memory to the OS, preventing unnecessary memory allocations. Some memory space for alignment is unused. Given that the sequence length is $n$, the memory loss is bounded by $(c-1)/n$.

By sharing common prefixes, the number of sequences that can be processed simultaneously is increased by approximately $1/(1-r)$. The sharing ratio $r$ is defined by the percentage of shared tokens $n_s/(n_p+n_c)$, and $n_c$ is the completion token count. In memory-limited inference scenarios, this helps increase the batch size and thus improve throughput.

The parent-child relationship defines the subset of sequences each chunk covers. The root node covers all sequences, and the leaf nodes cover only one. A key property of the prefix tree is that sequences covered by each chunk in the prefix tree are contiguous in the sequence index dimension. Therefore, slicing the query tensor in self-attention is particularly efficient during kernel computation, which will be discussed in more detail in the next section.

\subsection{Two-phase Partition (TPP)} \label{sec:TPP}

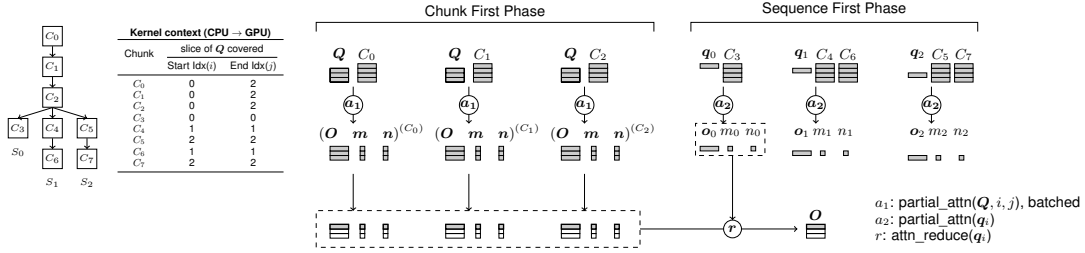
\begin{figure*}[htb]
  \centering
  \resizebox{0.9\linewidth}{!} {
      \begin{tikzpicture}[font=\sffamily\footnotesize]

\begin{scope}[xshift=0cm,yshift=4.0cm,
    chunk/.style = {align=center, inner sep=0pt, text centered, minimum size=1.1em, draw=black},
    chunkbold/.style = {align=center, inner sep=0pt, text centered, minimum size=1.1em, draw=black, line width=1.2pt}]

    \node [chunk] (0) {\tiny $C_0$};
    \node [chunk] (1) [below =0.6em of 0] {\tiny $C_1$};
    \node [chunk] (2) [below =0.6em of 1] {\tiny $C_2$};
    \node [chunk] (3) [below left =0.6em and 0.8em of 2] {\tiny $C_3$};
    \node [chunk] (4) [below =0.6em and 0.4em of 2] {\tiny $C_4$};
    \node [chunk] (5) [below right=0.6em and 0.8em of 2] {\tiny $C_5$};
    \node [chunk] (7) [below =0.6em of 4] {\tiny $C_6$};
    \node [chunk] (8) [below =0.6em of 5] {\tiny $C_7$};
    \node [draw=none] [below =0.2em of 3] {\tiny $S_0$};
    \node [draw=none] [below =0.2em of 7] {\tiny $S_1$};
    \node [draw=none] [below =0.2em of 8] {\tiny $S_2$};
    
    \draw [->] (0) -- (1);
    \draw [->] (1) -- (2);
    \draw [->] (2.south) -- (3.north);
    \draw [->] (2.south) -- (4.north);
    \draw [->] (2.south) -- (5.north);
    \draw [->] (4) -- (7);
    \draw [->] (5) -- (8);
\end{scope}

\begin{scope}[xshift=2.5cm,yshift=4cm, scale=0.7, transform shape]
    \node [shape=rectangle, align=center](table1) at (1.0,-2.0) {
            \textbf{Kernel context (CPU $\rightarrow$ GPU)} \\
            \begin{tabular}{ccc} 
                \toprule
                \multirow{2}{*}{Chunk} & \multicolumn{2}{c}{slice of $\bm{Q}$ covered }  \\ \cmidrule(lr){2-3}
                 & Start Idx($i$) & End Idx($j$) \\ \midrule
                $C_0$ & 0 & 2  \\
                $C_1$ & 0 & 2  \\
                $C_2$ & 0 & 2  \\
                $C_3$ & 0 & 0  \\
                $C_4$ & 1 & 1  \\
                $C_5$ & 2 & 2  \\
                $C_6$ & 1 & 1  \\
                $C_7$ & 2 & 2  \\
                \bottomrule
            \end{tabular}
        };
\end{scope}

\begin{scope}[xshift=20cm,yshift=0.0cm]
    \node [rectangle, align=left,] {
    $a_1$: partial\_attn($\bm{Q}, i, j$), batched \\
    $a_2$: partial\_attn($\bm{q}_i$) \\
    $r$: attn\_reduce($\bm{q}_i$)
    };
\end{scope}

\newcommand{\QC}[1]{%
    \node at (0.2, 0.6) {$\bm{Q}$};
    \draw[xstep=0.4,ystep=0.1,black,thin,fill=gray!40,] (0, 0) grid +(0.4, 0.3) rectangle (0, 0);
    \node at (0.8, 0.6) {$#1$};
    \foreach \y in {0,0.1,0.2,0.3} { \draw [fill=gray!40] (0.6,\y) rectangle ($(0.6,\y)+(0.4,0.1)$); }

    \draw[->] (0.5, -0.1) -- (0.5, -0.9);
    \node [circle,draw,fill=white,minimum size=6, inner sep=0] at (0.5, -0.5) {$\bm{a_1}$};
    
    \node at (0.9, -1.1) {$\left( \bm{O} \quad \bm{m} \quad \bm{n} \right) ^{(#1)}$};
    \foreach \y in {-1.7,-1.6,-1.5} { 
        \draw [fill=gray!40] (0.0,\y) rectangle +(0.4,0.1);
        \draw [fill=gray!40] (0.65,\y) rectangle +(0.1,0.1);
        \draw [fill=gray!40] (1.15,\y) rectangle +(0.1,0.1); }

    \draw[->] (0.5, -2) -- (0.5, -2.8);

    \foreach \y in {-3.2,-3.3,-3.4} { 
        \draw [] (0.0,\y) rectangle +(0.4,0.1);
        \draw [] (0.65,\y) rectangle +(0.1,0.1);
        \draw [] (1.15,\y) rectangle +(0.1,0.1); }
    \draw [fill=gray!40,thick] (0.0,-3.2) rectangle +(0.4,0.1);
    \draw [fill=gray!40,thick] (0.65,-3.2) rectangle +(0.1,0.1);
    \draw [fill=gray!40,thick] (1.15,-3.2) rectangle +(0.1,0.1);
}

\begin{scope}[xshift=6cm,yshift=3.0cm]
    \QC{C_0}
    \draw[-] (-0.3, 1.1) -- (-0.3, 1.3) -- (7, 1.3) node [midway, above, sloped] (TextNode) {Chunk First Phase} -- (7, 1.1);
    \draw [dashed] (-0.3, -3.6) rectangle +(7, 0.8);
\end{scope}

\begin{scope}[xshift=8.5cm,yshift=3.0cm]
    \QC{C_1}
\end{scope}

\begin{scope}[xshift=11cm,yshift=3.0cm]
    \QC{C_2}
\end{scope}

\begin{scope}[xshift=14cm,yshift=3.0cm]
    \node at (0.2, 0.6) {$\bm{q}_0$};
    \draw [fill=gray!40] (0,0.3) rectangle +(0.4,0.1);
    \node at (0.7, 0.6) {$C_3$};
    \foreach \y in {0,0.1,0.2,0.3} { \draw [fill=gray!40] (0.5,\y) rectangle +(0.4,0.1); }

    \draw[->] (0.5, -0.1) -- (0.5, -0.9);
    \node [circle,draw,fill=white,minimum size=6, inner sep=0] at (0.5, -0.5) {$\bm{a_2}$};

    \node at (0.2, -1.1) {$\bm{o}_0$};
    \foreach \y in {-1.5} { \draw [fill=gray!40] (0.0,\y) rectangle +(0.4,0.1); }
    \node at (0.65, -1.1) {$m_0$};
    \foreach \y in {-1.5} { \draw [fill=gray!40] (0.6,\y) rectangle +(0.1,0.1); }
    \node at (1.15, -1.1) {$n_0$};
    \foreach \y in {-1.5} { \draw [fill=gray!40] (1.1,\y) rectangle +(0.1,0.1); }
    
    \draw[-] (-0.2, 1.1) -- (-0.2, 1.3) -- (6, 1.3) node [midway, above, sloped] (TextNode) {Sequence First Phase} -- (6, 1.1);
\end{scope}

\begin{scope}[xshift=16cm,yshift=3.0cm]
    \node at (0.2, 0.6) {$\bm{q}_1$};
    \draw [fill=gray!40] (0,0.2) rectangle +(0.4,0.1);
    \node at (0.7, 0.6) {$C_4$};
    \foreach \y in {0,0.1,0.2,0.3} { \draw [fill=gray!40] (0.5,\y) rectangle +(0.4,0.1); }
    \node at (1.2, 0.6) {$C_6$};
    \foreach \y in {0,0.1,0.2,0.3} { \draw [fill=gray!40] (1.0,\y) rectangle +(0.4,0.1); }

    \draw[->] (0.5, -0.1) -- (0.5, -0.9);
    \node [circle,draw,fill=white,minimum size=6, inner sep=0] at (0.5, -0.5) {$\bm{a_2}$};
       
    \node at (0.2, -1.1) {$\bm{o}_1$};
    \foreach \y in {-1.6} { \draw [fill=gray!40] (0.0,\y) rectangle +(0.4,0.1); }
    \node at (0.65, -1.1) {$m_1$};
    \foreach \y in {-1.6} { \draw [fill=gray!40] (0.6,\y) rectangle +(0.1,0.1); }
    \node at (1.15, -1.1) {$n_1$};
    \foreach \y in {-1.6} { \draw [fill=gray!40] (1.1,\y) rectangle +(0.1,0.1); }

    \node at (0.5, -2.9) {$\bm{O}$};
    \draw [fill=gray!40,thick] (0.3,-3.2) rectangle +(0.4,0.1);
    \draw [] (0.3,-3.3) rectangle +(0.4,0.1);
    \draw [] (0.3,-3.4) rectangle +(0.4,0.1);

    \draw[dashed] (-2.1, -1.6) rectangle +(1.5, 0.7);
    \draw[->] (-3.3, -3.2) -- (0.1,-3.2);
    \node [circle,draw,fill=white,minimum size=11, inner sep=0] at (-1.3, -3.2) {$\bm{r}$};
    \draw[->] (-1.3, -1.6) -- (-1.3, -3);
\end{scope}

\begin{scope}[xshift=18.5cm,yshift=3.0cm]
    \node at (0.2, 0.6) {$\bm{q}_2$};
    \draw [fill=gray!40] (0,0.1) rectangle +(0.4,0.1);
    \node at (0.7, 0.6) {$C_5$};
    \foreach \y in {0,0.1,0.2,0.3} { \draw [fill=gray!40] (0.5,\y) rectangle +(0.4,0.1); }
    \node at (1.2, 0.6) {$C_7$};
    \foreach \y in {0,0.1,0.2,0.3} { \draw [fill=gray!40] (1.0,\y) rectangle +(0.4,0.1); }

    \draw[->] (0.5, -0.1) -- (0.5, -0.9);
    \node [circle,draw,fill=white,minimum size=6, inner sep=0] at (0.5, -0.5) {$\bm{a_2}$};

    \node at (0.2, -1.1) {$\bm{o}_2$};
    \foreach \y in {-1.7} { \draw [fill=gray!40] (0.0,\y) rectangle +(0.4,0.1); }
    \node at (0.65, -1.1) {$m_2$};
    \foreach \y in {-1.7} { \draw [fill=gray!40] (0.6,\y) rectangle +(0.1,0.1); }
    \node at (1.15, -1.1) {$n_2$};
    \foreach \y in {-1.7} { \draw [fill=gray!40] (1.1,\y) rectangle +(0.1,0.1); }
\end{scope}

\end{tikzpicture}
  }
  \caption{Two-phase partition kernel in ChunkAttention. The server is decoding sequences $S_0$, $S_1$, and $S_2$. They share chunks $C_0$, $C_1$ and $C_2$. In the chunk-first phase, queries $\bm{q}_0$, $\bm{q}_1$ and $\bm{q}_2$ are batched for self-attention with $C_0$, $C_1$ and $C_2$. Partial attention result $\bm{O}^{(C)}$, $\bm{m}^{(C)}$ and $\bm{n}^{(C)}$ are saved into memory. In the sequence-first phase, $\bm{o}_i$, $m_i$, and $n_i$ for each sequence are restored, and we continue processing the remaining chunks with respect to $\bm{q}_i$ only.}
  \label{fig:chunk_attn_kernel}
\end{figure*}

In this section, we dive into the self-attention kernel implementation on top of the unique prefix-aware KV cache storage.

During prefilling, we perform a prefix lookup to avoid repeated computation of KV projection and position embedding for matched prompt prefixes. For mismatched suffix tokens, KV projection and position embedding are still computed, and the key/value tensors are chunked and inserted into the prefix tree. Then we apply existing highly optimized self-attention kernels, \emph{e.g.}, FlashAttention \citep{dao2023flashattention}, on the entire key/value tensors.

During iterative decoding, self-attention is divided into chunk-first and sequence-first phases. The two phases focus on different slices of the query tensor, KV cache chunks, and use different parallelization strategies. The process is shown in Figure~\ref{fig:chunk_attn_kernel}. Since the head dimension is always partitioned, it is omitted and implicit in our discussion.

\textbf{Chunk-first Phase}. In the chunk-first phase, we only process chunks shared by multiple sequences. Since GPUs have more streaming multiprocessors (108 for A100) than the number of heads (32 for Llama 7B), and partitioning by heads under-utilizes hardware resources, we perform additional partition on keys/values. Chunking already provides convenience. The online softmax algorithm is adopted to avoid the synchronization requirement between partitions \citep{milakov2018online,dao2023flashattention}.

The computation is performed by traversing shared chunks in the prefix tree, executing the partial attention kernel \textit{partial\_attn} and saving the partial attention results into memory, as shown in Algorithm~\ref{alg:chunk_first}. The number of sequences (batch size) is denoted by $b$. $\bm{Q} \in \mathbb{R}^{b \times d}$ is the queries formed by concatenating the last token of all $b$ sequences in the latest decoding iteration.

\begin{algorithm}[hbt]
  \captionsetup{font=scriptsize}
  \scriptsize
  \caption{Self Attention: Chunk First (partition chunks)}\label{alg:chunk_first}
  \begin{algorithmic} 
    \Require{$\bm{Q} \in \mathbb{R}^{b \times d}$ (query), $T$(prefix tree)}
    \Ensure{$\bm{O} \in \mathbb{R}^{b \times d}$ (attention output)}
    \Function {AttnChunkFirst} {$\bm{Q}$, $T$}
      \State {Get chunks $C_1,...,C_k$ in $T$ that are shared by multiple sequences}
      \State {$\bm{O}, \bm{m}, \bm{n}$ $\gets$ $\bm{0}$, $\bm{0}$, $\bm{0}$}
      \For{$C$ $\gets$ $C_1$ to $C_k$}
        \State {$\bm{K}^{(C)}$, $\bm{V}^{(C)}$ $\gets$ key, value cache stored in $C$}
        \State {$i,j$ $\gets$ start index, end index of sequences covered by $C$}
        \State {$\bm{O}^{(C)}, \bm{m}^{(C)}, \bm{n}^{(C)}$ $\gets$ \textbf{\textit{partial\_attn}}($\bm{Q}$, $\bm{K}^{(C)}$, $\bm{V}^{(C)}$, $i$, $j$)}
        \State {Save partial attention result $\bm{O}^{(C)}, \bm{m}^{(C)}, \bm{n}^{(C)}$ to memory}
      \EndFor
    \EndFunction
  \end{algorithmic}
\end{algorithm}

The implementation of \textit{partial\_attn} is given by Eqn.~\eqref{eqn:kernel_map}. It computes the partial attention result $(\bm{O}, \bm{m}, \bm{n})^{(C)}$ with respect to each chunk $C$ independently, thus it can be parallelized. $\bm{Q}_{i:j,:}$ is a slice of $\bm{Q}$ for sequences ranging from $i$ to $j$ which share the KV cache stored in chunk $C$. The maximum attention weights vector $\bm{M}^{(C)}$ is the row-wise max over the last dimension of attention weights $\bm{W}^{(C)}$. The softmax normalization term $\bm{n}^{(C)}$ is the row-wise sum over the last dimension of $\bm{E}^{(C)}$. $\bm{M}^{(C)}$ and $\bm{n}^{(C)}$ are auxiliary variables introduced to further cumulate partial attention results of multiple chunks.

\begin{equation}
  \label{eqn:kernel_map}
  \resizebox{0.7\columnwidth}{!}{
  \begin{math}
  \begin{aligned}
    \bm{W}^{(C)} &= \bm{Q}_{i:j,:} \bm{K}^{(C)} \in \mathbb{R}^{(j-i) \times c} \\
    \bm{m}^{(C)} &= \textrm{max} \left( \bm{W}^{(C)} \right) \in \mathbb{R}^{(j-i)} \\
    \bm{E}^{(C)} &= \textrm{exp} \left( \bm{W}^{(C)} - \bm{m}^{(C)} \cdot \bm{1}^T \right) \in \mathbb{R}^{(j-i) \times c} \\
    \bm{n}^{(C)} &= \textrm{sum} \left( \bm{E}^{(C)} \right) \in \mathbb{R}^{(j-i)} \\
    \bm{O}^{(C)} &= \bm{E}^{(C)} \bm{V}^{(C)} \in \mathbb{R}^{(j-i) \times d}
  \end{aligned}
  \end{math}
  }
\end{equation}

The \textit{partial\_attn} efficiently accesses shared KV cache memory since self-attentions for multiple sequences are batched. The batching happens at a granularity of dot-product between queries $\bm{Q}_{i,:},...,\bm{Q}_{j,:}$ of sequences $S_{i},...,S_{j}$ and shared $\bm{K}^{(C)}$/$\bm{V}^{(C)}$. In addition to improved data locality, another advantage of batching is to turn the query from a vector into a matrix, allowing efficient matrix multiplications with tensor cores \citep{a100tensorcore}.

\textbf{Sequence-first Phase}. In the sequence-first phase, we load partial attention results of shared chunks from the chunk-first phase and continue processing chunks related to one specific sequence. We partition sequences, and each $\bm{q}$ handled by the sequence-first kernel is a vector by selecting the $i$-th row of $\bm{Q}$, as shown in Algorithm~\ref{alg:seq_first}.

\begin{algorithm}[hb]
  \captionsetup{font=scriptsize}
  \scriptsize
  \caption{Self Attention: Sequence First (partition sequences)}\label{alg:seq_first}
  \begin{algorithmic} 
    \Require{$\bm{Q} \in \mathbb{R}^{b \times d}$ (query), $T$(prefix tree)}
    \Ensure{$\bm{O} \in \mathbb{R}^{b \times d}$ (attention output)}
    \Function {AttnSeqFirst} {$\bm{Q}$, $T$}  
      \For{$\bm{q}$ $\gets$ $\bm{q}_1$ to $\bm{q}_b$}
        \State {$\bm{o}, m, n$ $\gets$ $\bm{0}$, $0$, $0$}
        \State {Get partial attn results $\left( \bm{O}, \bm{m}, \bm{n} \right)^{(C_1)},..., \left( \bm{O}, \bm{m}, \bm{n} \right)^{(C_k)}$}
        \For{$\left( \bm{O}, \bm{m}, \bm{n} \right)^{(C)}$ $\gets$ $\left( \bm{O}, \bm{m}, \bm{n} \right)^{(C_1)}$ to $\left( \bm{O}, \bm{m}, \bm{n} \right)^{(C_k)}$}
          \State {Partial attn of $\bm{q}$: $\left( \bm{o}, m, n \right)^{(C)}$ $\gets$ slicing $\left( \bm{O}, \bm{m}, \bm{n} \right)^{(C)}$}
          \State {\textbf{\textit{attn\_reduce}}($\bm{o}^{(C)}$, $m^{(C)}$, $n^{(C)}$, $\bm{o}$, $m$, $n$)}
        \EndFor        
        \State {Get chunks $C_{k+1},C_{k+2}...,C_{l}$ in $T$ with respect to $\bm{q}$ only}
        \For{$C$ $\gets$ $C_{k+1}$ to $C_l$}
          \State {$\bm{K}^{(C)}$, $\bm{V}^{(C)}$ $\gets$ key, value cache stored in $C$}
          \State {$i$ $\gets$ sequence index of $\bm{q}$}
          \State {$\left( \bm{o}, m, n \right)^{(C)}$ $\gets$ \textbf{\textit{partial\_attn}}($\bm{q}$, $\bm{K}^{(C)}$, $\bm{V}^{(C)}$, $i$, $i+1$)}
          \State {\textbf{\textit{attn\_reduce}}($\bm{o}^{(C)}$, $m^{(C)}$, $n^{(C)}$, $\bm{o}$, $m$, $n$)}
        \EndFor
      \EndFor
    \EndFunction
  \end{algorithmic}
\end{algorithm}

The \textit{attn\_reduce} repeatly merges partial attention result of one chunk $(\bm{o}, m, n)^{(C)}$ produced by the \textit{partial\_attn} into the cumulative attention result $(\bm{o}, m, n)$ by scaling them with $x^{(C)}$ and $y^{(C)}$ respectively. Eqn.~\eqref{eqn:kernel_reduce} shows the process. $\bm{O}_{i,:}$, $\bm{m}_{i}$ and $\bm{n}_{i}$ are slices for sequence of index $i$. The final attention output is given by $\bm{O}/\bm{n}$ element-wise. 

The sequence-first phase is efficient in concurrency since \textit{partial\_attn} and \textit{attn\_reduce} are performed locally, without communication between thread blocks. However, without the partial attention results generated by the chunk-first phase, it needs to load shared chunks from RAM $b$ times, which adds significant MOPs. The two-phase partition algorithm balances parallelization and cache locality.

\begin{equation}
  \label{eqn:kernel_reduce}
  \resizebox{0.7\columnwidth}{!}{
  \begin{math}
  \begin{aligned}
    x^{(C)} &= \textrm{exp} \left( m^{(C)} - \textrm{max} \left( m^{(C)}, \bm{m}_{i} \right) \right) \in \mathbb{R} \\
    y^{(C)} &= \textrm{exp} \left( \bm{m}_{i} - \textrm{max} \left( m^{(C)}, \bm{m}_{i} \right) \right) \in \mathbb{R} \\
    \bm{O}_{i,:} &= x^{(C)} \bm{o}^{(C)} + y^{(C)} \bm{O}_{i,:} \in \mathbb{R}^{d} \\
    \bm{n}_{i} &= x^{(C)} n^{(C)} + y^{(C)} \bm{n}_{i} \in \mathbb{R} \\
    \bm{m}_{i} &= \textrm{max} \left( m^{(C)}, \bm{m}_{i}\right) \in \mathbb{R} \\
  \end{aligned}
  \end{math}
  }
\end{equation}

\subsection{Further Optimizations}

The prefix tree structure is maintained in CPU memory. To run the two-phase partition kernel on GPU, we must generate certain context from the prefix tree, including the chunk $C$, the start index $i$ and end index $j$ of its covered sequences, and copy the context ($C$, $i$, $j$) from CPU to GPU memory. For example, in Figure~\ref{fig:chunk_attn_kernel}, we need to generate and copy $(C_0/C_1/C_2, 0, 2)$, $(C_3, 0, 0)$, $(C_4/C_6, 1, 1)$, and $(C_5/C_7, 2, 2)$. ChunkAttention manages the overhead in two ways: i) latency hiding. The context generation step on CPU can be overlapped with other kernels on GPU prior to self-attention. ii) lazy context copy. The prefix tree does not change at every decoding iteration. We can cache the context in GPU memory and only trigger memory copy when the tree structure changes. Triggers are chunk full for every $c$ iterations, new sequence joining, and completed sequence leaving. The overhead is amortized.

The temporary memory allocated for partial attention results in the chunk-first phase can be eliminated by executing \textit{attn\_reduce} right after \textit{partial\_attn} to directly merge partial attention results into the final result. Since multiple shared chunks with a parent-child relationship in the prefix tree write into the same slice of $(\bm{O},\bm{m},\bm{n})$, \textit{attn\_reduce} needs to be serialized. On GPU devices, atomic operations are heavy, and we do not use this approach. However, on CPU devices, the overhead of serializing is insignificant, and reduction can be implemented using spin locks.

\section{Experiments} \label{sec:exp}

The evaluations are conducted at both the self-attention microkernel level and the end-to-end GPT-style model level. The microkernel level evaluations only capture time spent in the self-attention CUDA kernel. The side effects of PAKV and TPP, \emph{e.g.}, prefix tree operations, are captured in end-to-end evaluations. We run all experiments with NVIDIA A100 GPU (80G) and CUDA 11.8.

\subsection{Microkernel Evaluation}

\noindent\textbf{Baselines}. We select four self-attention implementations as baselines: Naive PyTorch implementation by the formula $\textrm{softmax} (\bm{Q} \bm{K}^T/\sqrt{d})\bm{V}$, the memory-efficient self-attention implemented in xformers \citep{xFormers2022}, FlashAttention integrated in PyTorch \citep{dao2022flashattention}, and PagedAttention in vLLM \citep{vllm}.

Since Naive, xformers, and FlashAttn are all built on monolithic KV tensors, they cannot be prefix-aware by partially sharing KV cache of prompt prefixes. PagedAttn does not implement PAKV either. However, its paging design enables us to manually create a fixed page table, mapping virtually non-shared memory to the same physical memory. It simulates the KV cache sharing scenario and helps us observe the performance of PagedAttn's CUDA kernel, which is denoted as PagedAttn*. None of the kernels support the TPP algorithm.

\noindent\textbf{Workload}. Sequences are processed in batch mode, and the batch size is $b$. All sequences within the same batch start and finish simultaneously. Each sequence is prefilled with $n_{p}$ prompt tokens, and the leading $n_s$ tokens are common prefixes. The task is to decode the next $n_{c}$ completion tokens iteratively. We measure the decoding latency $t$ and the throughput defined by token rate(tokens per second or TPS, $n_c*b/t$). For all experiments, the head dimension $d$ is 128, the number of heads $h$ is 32, and the chunk size $c$ is 64. All tensors are in FP16.


\noindent\textbf{Results}. We run experiments to observe the performance gain brought by PAKV and TPP by varying the following system hyperparameters: prompt and shared token count, completion token count, and batch size.

Table~\ref{tab:kernel_latency} shows the latency of self-attention implementations given various prompt and shared token counts. ChunkAttn and PagedAttn* outperform Naive, xformers, FlashAttn, and PagedAttn, which are agnostic to shared token count. The Naive is 6.6$\times$ and 2.1$\times$ slower than ChunkAttn and PagedAttn*, respectively ($n_s$=4096). By comparing PagedAttn* and PagedAttn, we observe the performance gain brought by sharing KV cache memory physically. Although PagedAttn* does not implement PAKV or TPP, the hardware cache helps reduce its latency by up to 52\% compared to PagedAttn ($n_s$=4096): repeatedly accessing the same physical memory blocks provides significant performance gain. The benefit of TPP can be further seen by comparing PagedAttn* and ChunkAttn. ChunkAttn outperforms PagedAttn* by 2.8-3.2$\times$, with a range of $n_s$ from 1024 to 4096. TPP does not cause performance regression when no token is shared ($n_s$=0, ChunkAttn vs. PagedAttn* in Table~\ref{tab:kernel_latency}). As a result, TPP should always be enabled.

\begin{figure}[tb]
  \centering
  \begin{minipage}{0.54\linewidth}
    \centering
    \includegraphics[width=1.0\linewidth]{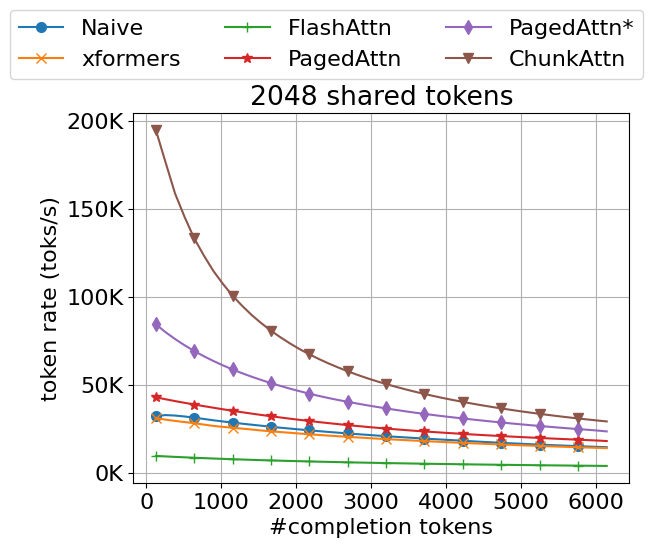}
  \end{minipage}
  \begin{minipage}{0.44\linewidth}
  \centering
  \resizebox{\columnwidth}{!}{
    \begin{tabular}{ccrrr}
    \toprule
    \multirow{2}{*}{\textbf{$n_s$}} & \multirow{2}{*}{\textbf{$n_c$}} & \multicolumn{3}{c}{Token Rate($\times 10^3$) (toks/s)} \\ \cmidrule(lr){3-5}
    & & \multicolumn{1}{c}{Paged} & \multicolumn{1}{c}{Chunk} & \multicolumn{1}{c}{Speedup} \\ \midrule
    \multirow{4}{*}{1024} & {256} & {76.35} & {241.93} & \textbf{3.2 $\times$} \\
    & {512} & {69.15} & {186.44} & {2.7 $\times$} \\
    & {1024} & {58.12} & {127.85} & {2.2 $\times$} \\
    \midrule
    \multirow{4}{*}{2048} & {512} & {39.85} & {145.41} & \textbf{3.6 $\times$} \\
    & {1024} & {36.18} & {107.37} & {3.0 $\times$} \\
    & {2048} & {30.17} & {70.33} & {2.3 $\times$} \\
    \midrule
    \multirow{4}{*}{4096} & {512} & {21.04} & {101.69} & \textbf{4.8 $\times$} \\
    & {1024} & {19.85} & {81.69} & {4.1 $\times$} \\
    & {2048} & {17.98} & {58.33} & {3.2 $\times$} \\
    & {4096} & {15.12} & {37.05} & {2.4 $\times$} \\
    \bottomrule
    \end{tabular}
  }
  \end{minipage}
  \caption{Throughput in token rate when generating up to $n_c$ completion tokens, given $n_s$ prefix tokens are shared. Chunk size $c$=64, batch size $b$=32.}
  \label{fig:cuda_attn_tps}
\end{figure}

As the decoding proceeds, sequences start to diverge, and the performance gain of ChunkAttn gradually decreases, as shown in Figure~\ref{fig:cuda_attn_tps}. Given 2048 shared tokens, ChunkAttn yields 3.6$\times$ token rate improvement compared to PagedAttn when $n_c$ reaches 512, and the speedup drops to 2.3$\times$ when $n_c$ reaches 2048. However, it is still a significant improvement. The improvement of ChunkAttn over PagedAttn* is lower since PagedAttn* benefits from physically shared KV cache memory, and only TPP makes a difference here. However, given $n_s$=2048, ChunkAttn is still 2.0$\times$ (145K against 73K) and 1.5$\times$ (70K against 46K) faster than PagedAttn* when $n_c$ reaches 512 and 2048 respectively.

\begin{table}
  \centering
  \resizebox{\columnwidth}{!}{
  \begin{tabular}{lcrrrrrr}
  \toprule
  \multirow{2}{*}{\textbf{$n_p$}} & \multirow{2}{*}{\textbf{$n_s$}} & \multicolumn{6}{c}{Latency (\textmu s)} \\ \cmidrule(lr){3-8}
  & & Naive & xformers & FlashAttn & PagedAttn & PagedAttn* & ChunkAttn \\
  \midrule
  {1024} & {0} & {363.35} & {378.19} & {1586.73} & {356.17} & {355.82} & {332.50} \\ 
  {1024} & {512} & {364.73} & {385.79} & {1587.14} & {355.88} & {257.74} & \textbf{198.87} \\ 
  {1024} & {768} & {362.43} & {378.50} & {1591.61} & {356.02} & {215.18} & \textbf{131.21} \\ 
  {1024} & {1024} & {361.76} & {379.36} & {1586.90} & {355.44} & {154.46} & \textbf{56.00} \\ 
  \midrule
  {2048} & {0} & {686.40} & {816.44} & {3175.25} & {702.98} & {703.50} & {655.44} \\ 
  {2048} & {1024} & {687.52} & {828.76} & {3173.53} & {703.35} & {505.32} & \textbf{384.37} \\ 
  {2048} & {1536} & {685.78} & {820.19} & {3174.96} & {702.90} & {421.25} & \textbf{247.14} \\ 
  {2048} & {2048} & {688.41} & {823.60} & {3152.25} & {703.72} & {338.41} & \textbf{110.48} \\ 
  \midrule
  {4096} & {0} & {1369.52} & {1720.00} & {6289.55} & {1400.61} & {1400.17} & {1301.78} \\ 
  {4096} & {2048} & {1370.47} & {1722.42} & {6303.21} & {1400.99} & {998.78} & \textbf{747.56} \\ 
  {4096} & {3072} & {1369.74} & {1725.57} & {6301.41} & {1400.30} & {828.98} & \textbf{477.66} \\ 
  {4096} & {4096} & {1370.41} & {1713.13} & {6300.65} & {1399.51} & {663.84} & \textbf{206.22} \\ 
  \bottomrule
  \end{tabular}
  }
  \caption{Latency of self-attention kernel given $n_p$ context tokens and $n_s$ prefix tokens are shared. Chunk size $c$=64, batch size $b$=32.}
  \label{tab:kernel_latency}
\end{table}

\begin{figure}[tb]
  \centering
  \begin{subfigure}{0.48\columnwidth}
    \centering
    \includegraphics[width=1.0\linewidth]{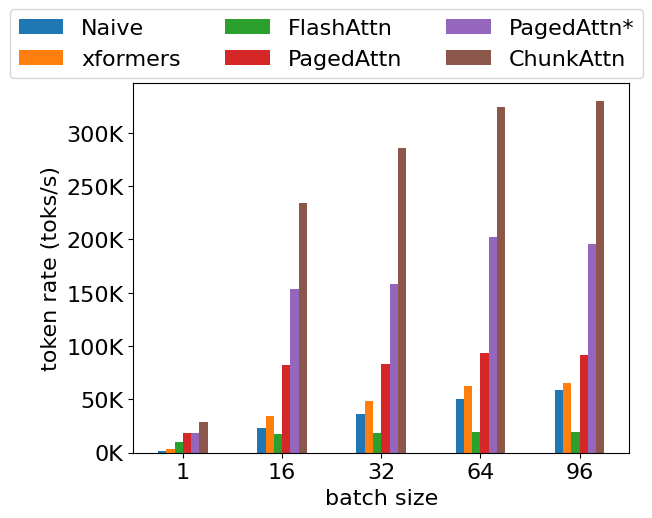}
    \caption{$n_s$=1024}
  \end{subfigure}
  \hfill
  \begin{subfigure}{0.48\linewidth}
    \centering
    \includegraphics[width=1.0\linewidth]{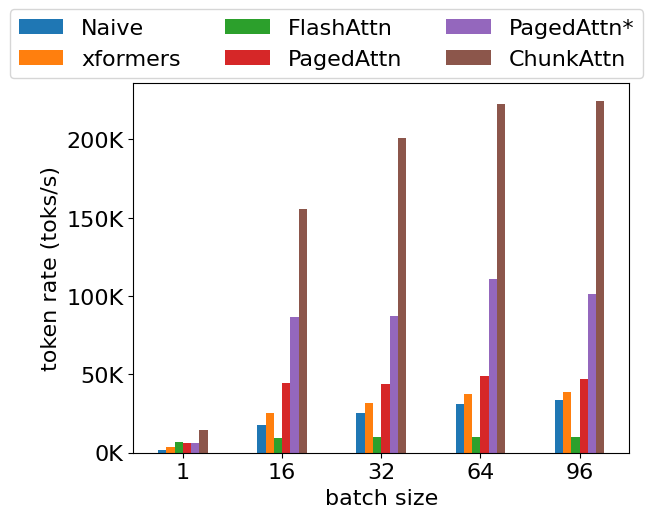}
    \caption{$n_s$=2048}
  \end{subfigure}
  \caption{Token rate when decoding up to $n_c$=64 completion tokens given various batch sizes. Chunk size $c$=64.}
  \label{fig:cuda_tps_batch}
\end{figure}

Figure~\ref{fig:cuda_tps_batch} focuses on varying the batch size. For all implementations except ChunkAttn and PagedAttn*, the throughput peaks when the batch size reaches 16 due to memory-bound. 
Given $n_s$ is 2048, ChunkAttn's throughput continues to grow from 155K to 224K toks/s for the batch size ranging from 16 to 96 due to better data locality and improved arithmetic intensity.

\subsection{End-to-end Evaluation}

ChunkLlama is built on top of Huggingface Llama and vLLM's optimized kernels (layer normalization and rotary embedding) under Apache-2.0 license, but the attention module is substituted by ChunkAttn. We run all experiments on the Open Llama2 7B model in FP16 \cite{openlm2023openllama,together2023redpajama,touvron2023llama}.

\noindent\textbf{Baselines}. We select two widely used and optimized LLMs serving toolkits with proven production usages: the start-of-the-art vLLM 0.2.7 \citep{vllm} and Huggingface's Text Generation Inference (TGI) 1.3.4 \cite{tgi}. 

\noindent\textbf{Workload}. Requests arrive at the server randomly following the Poisson arrival process \citep{HILL199249} parameterized by $\lambda$, which is the average requests per second (RPS). The actual batch size is adjusted dynamically by each system during decoding, and we configure its maximum to 32 equally. Application developers provide no information about the shared prompt prefix for the service provider to pre-configure. We measure the normalized latency (ms/tok or 1/TPS) as in vLLM, which is the mean of each request's end-to-end latency $t$ (including queuing time) divided by the completion token count $n_c$, and the peak memory bytes used by KV cache.

\begin{figure}[htb]
  \centering
  \begin{subfigure}{0.85\columnwidth}
    \centering
    \includegraphics[width=1.0\linewidth]{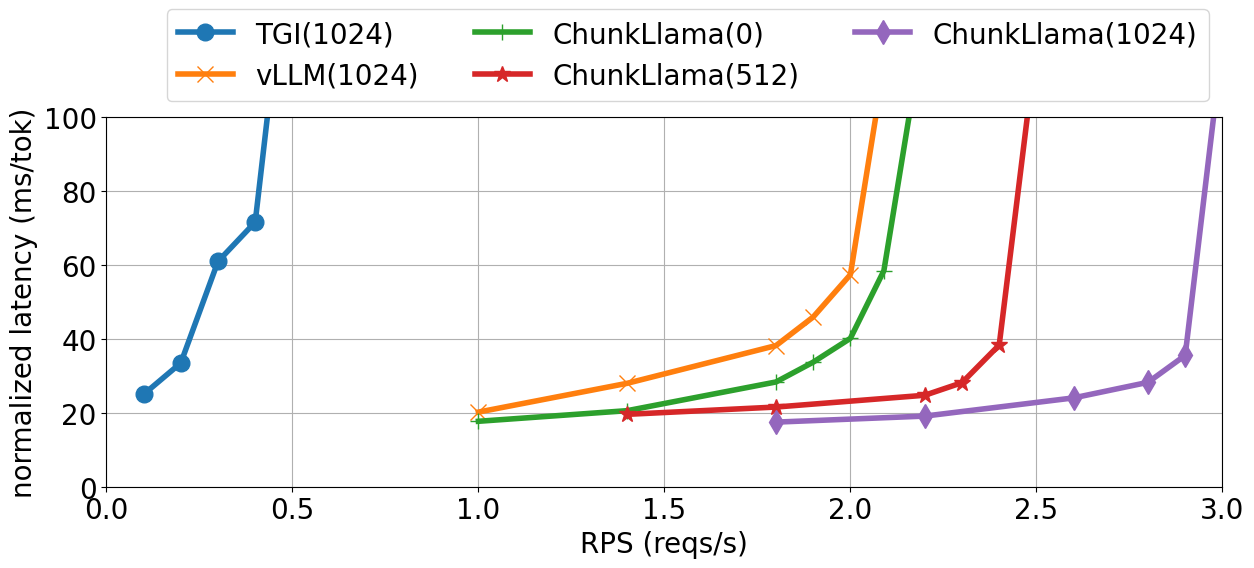}
    \caption{$n_p$=1024}
  \end{subfigure}
  \hfill
  \begin{subfigure}{0.85\linewidth}
    \centering
    \includegraphics[width=1.0\linewidth]{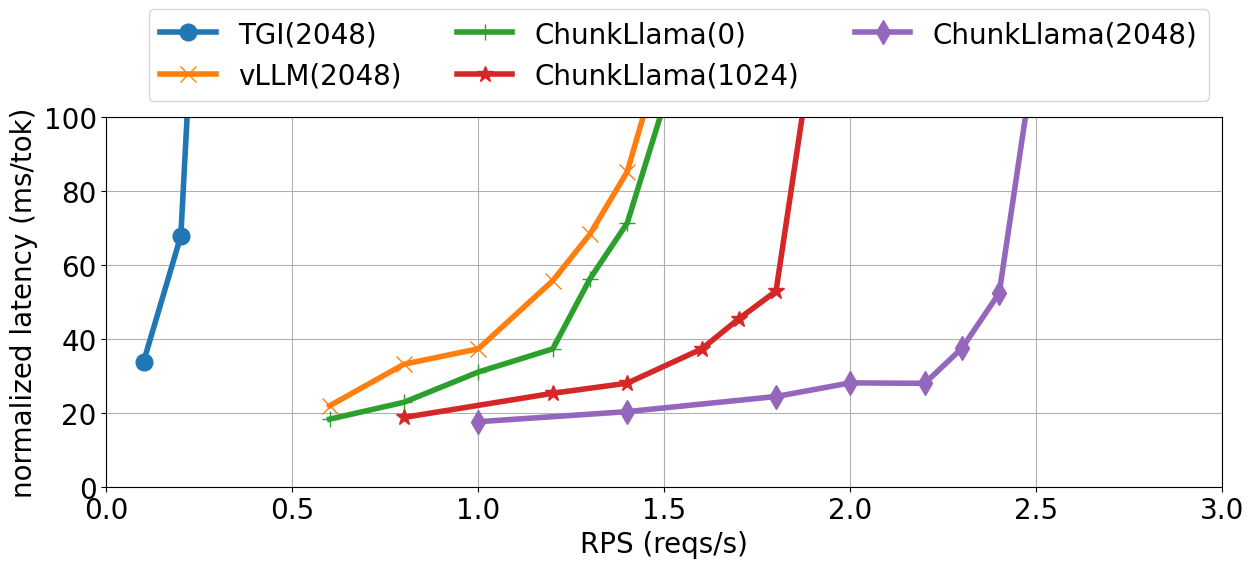}
    \caption{$n_p$=2048}
  \end{subfigure}
  \caption{Normalized latency given different request arrival rates (RPS). Each line is marked by the system and shared prompt token count: system($n_s$).}
  \label{fig:e2e_latency_vs_rps}
\end{figure}

\noindent\textbf{Results}. ChunkLlama yields the fastest inference speed, as shown in Figure~\ref{fig:e2e_latency_vs_rps}. ChunkLlama can achieve 1.6$\times$ (2.9 against 1.8) and 2.3$\times$ (2.3 against 1.0) higher throughput compared to vLLM when 1024 and 2048 prefix tokens are shared while maintaining a normalized latency of less than 40 ms/token. Table~\ref{tab:e2e_latency} compares the latency and KV cache memory usage of our ChunkLlama to vLLM. No performance regression is observed in ChunkLlama without shared prefix tokens. The KV cache memory usage is reduced by 70\%-90\% with long shared prefixes. The peak batch size is also reduced by 20\%-40\% since ChunkLlama can decode faster. 

\begin{table}[htb]
  \centering
  \resizebox{\columnwidth}{!}{
  \begin{tabular}{llllrrrrrr}
  \toprule
  \multirow{2}{*}{\textbf{$n_p$}} & \multirow{2}{*}{\textbf{$n_s$}} & \multirow{2}{*}{\textbf{$n_c$}} & \multirow{2}{*}{RPS} & \multicolumn{2}{c}{Latency (ms/tok)} & \multicolumn{2}{c}{Peak KV Cache (GB)} & \multicolumn{2}{c}{Peak Batch Size} \\ \cmidrule(lr){5-6} \cmidrule(lr){7-8} \cmidrule(lr){9-10}
  & & & & vLLM & ChunkLlama & vLLM & ChunkLlama & vLLM & ChunkLlama \\
  \midrule
  {1024} & {0} & {512} &  1.0 & {19.92} & {19.11} & {14.73} & {11.90} & 23 & 18\\
  {1024} & {1024} & {512} & 1.0 & {20.80} & {\textbf{14.07}} & {14.79} & {\textbf{3.28}} & 23 & 14 \\
  \midrule
  {2048} & {0} & {512} & 0.6 & {21.90} & {19.43} & {21.70} & {22.41} & 19 & 20 \\
  {2048} & {2048} & {512} & 0.6 & {21.61} & {\textbf{15.20}} & {21.09} & {\textbf{3.40}} & 19 & 12 \\
  \midrule
  {4096} & {0} & {512} & 0.4 & {26.23} & {26.88} & {34.59} & {35.13} & 16 & 16 \\
  {4096} & {4096} & {512} & 0.4 & {27.62} & {\textbf{17.16}} & {35.42} & {\textbf{4.00}} & 16 & 11 \\
  \bottomrule
  \end{tabular}
  }
  \caption{Normalized latency, peak KV cache memory, and batch size reached during decoding.}
  \label{tab:e2e_latency}
\end{table}

\section{Related Work}

The most relevant work on optimizing the memory utilization of KV cache is PagedAttention in vLLM \citep{vllm}. It introduces the paging technique in OS to solve the problem of memory waste caused by dynamic and unknown sequence lengths during decoding. However, only a proposal on service providers to pre-configure shared prompts is mentioned, and it is not implemented in vLLM releases (up to 0.2.7). Our solution, which differs from the paging one, uses the prefix tree to manage memory and aims to discover redundancy in KV cache across user requests at runtime automatically. The solution is more practical for multi-tenant deployment scenarios where service providers centrally host models and have requirements on scalability. According to vLLM, the shared KV cache is similar to the dynamic link library shared by multiple processes. vLLM's strategy is to compile before publishing (AoT). We expect to compile in real-time (JIT). Based on the context captured in the prefix tree, our work further proposes a two-phase partition algorithm to explore the optimization opportunities shared system prompts bring to the self-attention kernel, which is another difference between our work and the existing work.

Partition strategies in ChunkAttention are built on online softmax \citep{milakov2018online} and inspired by FlashAttention \citep{dao2022flashattention,dao2023flashattention}, which adopted the same algorithm. FlashAttention thoroughly researched and implemented various tiling techniques, accelerating self-attention by 2-4$\times$ while cutting memory operations by 10-20$\times$. FlashAttention-2 altered tiling strategies and additionally doubled the speed. However, FlashAttention is inflexible regarding non-contiguous memory or variable sequence lengths, making it more suitable for model training than inference. There is little gain when the query token count is always one during decoding. ChunkAttention handles variable sequence lengths during decoding and batches attention operations of several sequences to reduce memory operations. As a result, our work and FlashAttention are complementary.

\section{Conclusion}

In this paper, we propose ChunkAttention, a novel self-attention module, to efficiently manage KV cache and accelerate the self-attention kernel for LLMs inference. We successfully adopt the prefix tree to create a prefix-aware KV cache. It addresses the challenge of detecting and removing redundant KV cache at runtime. We evaluate ChunkAttention in various configurations and at different levels, proving its feasibility and the side effects can be managed. Experiments show that the ChunkAttention kernel can achieve comparable throughput with SOTA PagedAttention kernel without shared system prompts and can outperform it by 3.2-4.8$\times$ with a shared system prompt of 1024 to 4096 tokens on A100 (80G) by applying prefix-aware KV cache and two-phase partition.
\section{Limitations} \label{sec:limitations}

\noindent\textbf{The Position of System Prompt.} To share the key/value tensors in memory, the shared system prompt must appear at the beginning of the sequence. Although this is the most common practice in many works and systems \citep{lu2023chameleon,qian2023creator,saadfalcon2023pdftriage,zhuang2023toolqa}, it is not mandatory. \citet{liu2023lost} reveals that language model performance degrades significantly when changing the position of relevant information, indicating that models struggle to access and use information in long input contexts robustly. In particular, performance is often lowest when models must use information in the middle of long input contexts. As a result, when application developers do not put the system prompt at the beginning for performance concerns after evaluations or unintentional mistakes, KV caches of the entire sequences are different, and PAKV cannot save memory in this case, although they have a large number of tokens in common.

\noindent\textbf{Fine Tuning.} In addition to using system prompts, fine-tuning is another promising way to infuse domain knowledge into LLMs \citep{houlsby2019parameter,hu2023llmadapters}. Due to the high training and deployment cost, LLMs are typically pre-trained and centrally hosted for multiple applications to share. It is not cost-efficient for each application to fine-tune models and deploy private instances. However, fine-tuning may become more practical and popular as hardware and software environments evolve. In this case, we no longer need to design long system prompts for each application, and the sharing opportunities of system prompts are reduced. As of today, we have not seen promising and cost-efficient fine-tuning and hosting solutions in this direction than using system prompts.

\noindent\textbf{Model and Hardware Compatibility.} To achieve the best performance, ChunkAttention implements the two-phase partition kernel with the low-level CUDA programming instead of leveraging high-level primitives in cuDNN \citep{oneapisr13:online} or PyTorch. We tune its performance for the most common LLM configurations, \emph{e.g.}, 128 head dimension size, and hardware, \emph{e.g.}, NVIDIA A100, GeForce RTX 4090, and Intel Xeon CPU. For other configurations and hardware, we need to tune and verify the performance case by case, which adds significant development costs. We believe community efforts are needed to generalize the two-phase partition algorithm and make it compatible with more model configurations and hardware.

\bibliography{custom}

\begin{thebibliography}{51}
\expandafter\ifx\csname natexlab\endcsname\relax\def\natexlab#1{#1}\fi

\bibitem[{Aminabadi et~al.(2022)Aminabadi, Rajbhandari, Awan, Li, Li, Zheng,
  Ruwase, Smith, Zhang, Rasley, and He}]{10046087}
Reza~Yazdani Aminabadi, Samyam Rajbhandari, Ammar~Ahmad Awan, Cheng Li, Du~Li,
  Elton Zheng, Olatunji Ruwase, Shaden Smith, Minjia Zhang, Jeff Rasley, and
  Yuxiong He. 2022.
\newblock \href {https://doi.org/10.1109/SC41404.2022.00051} {Deepspeed
  inference: Enabling efficient inference of transformer models at
  unprecedented scale}.
\newblock In \emph{SC22: International Conference for High Performance
  Computing, Networking, Storage and Analysis}, pages 1--15.

\bibitem[{Anil et~al.(2023)Anil, Dai, Firat, Johnson, Lepikhin, Passos,
  Shakeri, Taropa, Bailey, Chen et~al.}]{palm2}
Rohan Anil, Andrew~M Dai, Orhan Firat, Melvin Johnson, Dmitry Lepikhin,
  Alexandre Passos, Siamak Shakeri, Emanuel Taropa, Paige Bailey, Zhifeng Chen,
  et~al. 2023.
\newblock Palm 2 technical report.
\newblock \emph{arXiv e-prints}, pages arXiv--2305.

\bibitem[{Anthropic(2023)}]{anthropic:online}
Anthropic. 2023.
\newblock How to use system prompts.
\newblock
  \url{https://docs.anthropic.com/claude/docs/how-to-use-system-prompts}.

\bibitem[{Brown et~al.(2020)Brown, Mann, Ryder, Subbiah, Kaplan, Dhariwal,
  Neelakantan, Shyam, Sastry, Askell et~al.}]{gpt3}
Tom Brown, Benjamin Mann, Nick Ryder, Melanie Subbiah, Jared~D Kaplan, Prafulla
  Dhariwal, Arvind Neelakantan, Pranav Shyam, Girish Sastry, Amanda Askell,
  et~al. 2020.
\newblock Language models are few-shot learners.
\newblock \emph{Advances in neural information processing systems},
  33:1877--1901.

\bibitem[{Chang et~al.(2023)Chang, Wang, Wang, Wu, Yang, Zhu, Chen, Yi, Wang,
  Wang, Ye, Zhang, Chang, Yu, Yang, and Xie}]{chang2023survey}
Yupeng Chang, Xu~Wang, Jindong Wang, Yuan Wu, Linyi Yang, Kaijie Zhu, Hao Chen,
  Xiaoyuan Yi, Cunxiang Wang, Yidong Wang, Wei Ye, Yue Zhang, Yi~Chang,
  Philip~S. Yu, Qiang Yang, and Xing Xie. 2023.
\newblock \href {http://arxiv.org/abs/2307.03109} {A survey on evaluation of
  large language models}.

\bibitem[{Choquette et~al.(2021)Choquette, Gandhi, Giroux, Stam, and
  Krashinsky}]{a100tensorcore}
Jack Choquette, Wishwesh Gandhi, Olivier Giroux, Nick Stam, and Ronny
  Krashinsky. 2021.
\newblock \href {https://doi.org/10.1109/MM.2021.3061394} {Nvidia a100 tensor
  core gpu: Performance and innovation}.
\newblock \emph{IEEE Micro}, 41(2):29--35.

\bibitem[{Chu et~al.(2023)Chu, Chen, Chen, Yu, He, Wang, Peng, Liu, Qin, and
  Liu}]{cotsurvey}
Zheng Chu, Jingchang Chen, Qianglong Chen, Weijiang Yu, Tao He, Haotian Wang,
  Weihua Peng, Ming Liu, Bing Qin, and Ting Liu. 2023.
\newblock \href {http://arxiv.org/abs/2309.15402} {A survey of chain of thought
  reasoning: Advances, frontiers and future}.

\bibitem[{Computer(2023)}]{together2023redpajama}
Together Computer. 2023.
\newblock \href {https://github.com/togethercomputer/RedPajama-Data}
  {Redpajama-data: An open source recipe to reproduce llama training dataset}.

\bibitem[{Dao(2023)}]{dao2023flashattention}
Tri Dao. 2023.
\newblock Flashattention-2: Faster attention with better parallelism and work
  partitioning.
\newblock \emph{arXiv preprint arXiv:2307.08691}.

\bibitem[{Dao et~al.(2022)Dao, Fu, Ermon, Rudra, and
  R{\'e}}]{dao2022flashattention}
Tri Dao, Dan Fu, Stefano Ermon, Atri Rudra, and Christopher R{\'e}. 2022.
\newblock Flashattention: Fast and memory-efficient exact attention with
  io-awareness.
\newblock \emph{Advances in Neural Information Processing Systems},
  35:16344--16359.

\bibitem[{Dong et~al.(2023)Dong, Li, Dai, Zheng, Wu, Chang, Sun, Xu, Li, and
  Sui}]{iclsurvey}
Qingxiu Dong, Lei Li, Damai Dai, Ce~Zheng, Zhiyong Wu, Baobao Chang, Xu~Sun,
  Jingjing Xu, Lei Li, and Zhifang Sui. 2023.
\newblock \href {http://arxiv.org/abs/2301.00234} {A survey on in-context
  learning}.

\bibitem[{Gao et~al.(2018)Gao, Yu, Wu, and Li}]{gao2018low}
Pin Gao, Lingfan Yu, Yongwei Wu, and Jinyang Li. 2018.
\newblock Low latency rnn inference with cellular batching.
\newblock In \emph{Proceedings of the Thirteenth EuroSys Conference}, pages
  1--15.

\bibitem[{Gemini(2023)}]{geminiteam2023gemini}
Gemini. 2023.
\newblock \href {http://arxiv.org/abs/2312.11805} {Gemini: A family of highly
  capable multimodal models}.

\bibitem[{Geng and Liu(2023)}]{openlm2023openllama}
Xinyang Geng and Hao Liu. 2023.
\newblock \href {https://github.com/openlm-research/open_llama} {Openllama: An
  open reproduction of llama}.

\bibitem[{gyudoza(2023)}]{leakedsystemprompts}
gyudoza. 2023.
\newblock jujumilk3/leaked-system-prompts: Collection of leaked system prompts.
\newblock \url{https://github.com/jujumilk3/leaked-system-prompts}.

\bibitem[{Hill(1992)}]{HILL199249}
Steve Hill. 1992.
\newblock \href
  {https://doi.org/https://doi.org/10.1016/B978-0-08-050755-2.50022-1} {A
  simple fast memory allocator}.
\newblock In DAVID KIRK, editor, \emph{Graphics Gems III (IBM Version)}, pages
  49--50. Morgan Kaufmann, San Francisco.

\bibitem[{Houlsby et~al.(2019)Houlsby, Giurgiu, Jastrzebski, Morrone,
  De~Laroussilhe, Gesmundo, Attariyan, and Gelly}]{houlsby2019parameter}
Neil Houlsby, Andrei Giurgiu, Stanislaw Jastrzebski, Bruna Morrone, Quentin
  De~Laroussilhe, Andrea Gesmundo, Mona Attariyan, and Sylvain Gelly. 2019.
\newblock Parameter-efficient transfer learning for nlp.
\newblock In \emph{International Conference on Machine Learning}, pages
  2790--2799. PMLR.

\bibitem[{Hu et~al.(2023)Hu, Wang, Lan, Xu, Lim, Bing, Xu, Poria, and
  Lee}]{hu2023llmadapters}
Zhiqiang Hu, Lei Wang, Yihuai Lan, Wanyu Xu, Ee-Peng Lim, Lidong Bing, Xing Xu,
  Soujanya Poria, and Roy Ka-Wei Lee. 2023.
\newblock \href {http://arxiv.org/abs/2304.01933} {Llm-adapters: An adapter
  family for parameter-efficient fine-tuning of large language models}.

\bibitem[{HuggingFace(2023)}]{tgi}
HuggingFace. 2023.
\newblock huggingface/text-generation-inference: Large language model text
  generation inference.
\newblock \url{https://github.com/huggingface/text-generation-inference}.

\bibitem[{Jin et~al.(2023)Jin, Wu, Brooks, and Wei}]{jin2023s}
Yunho Jin, Chun-Feng Wu, David Brooks, and Gu-Yeon Wei. 2023.
\newblock S3: Increasing gpu utilization during generative inference for higher
  throughput.
\newblock \emph{arXiv preprint arXiv:2306.06000}.

\bibitem[{Kim et~al.(2023)Kim, Hooper, Wattanawong, Kang, Yan, Genc, Dinh,
  Huang, Keutzer, Mahoney, Shao, and Gholami}]{kim2023stack}
Sehoon Kim, Coleman Hooper, Thanakul Wattanawong, Minwoo Kang, Ruohan Yan,
  Hasan Genc, Grace Dinh, Qijing Huang, Kurt Keutzer, Michael~W. Mahoney,
  Yakun~Sophia Shao, and Amir Gholami. 2023.
\newblock \href {http://arxiv.org/abs/2302.14017} {Full stack optimization of
  transformer inference: a survey}.

\bibitem[{Kwon et~al.(2023)Kwon, Li, Zhuang, Sheng, Zheng, Yu, Gonzalez, Zhang,
  and Stoica}]{vllm}
Woosuk Kwon, Zhuohan Li, Siyuan Zhuang, Ying Sheng, Lianmin Zheng, Cody~Hao Yu,
  Joseph~E. Gonzalez, Hao Zhang, and Ion Stoica. 2023.
\newblock \href {http://arxiv.org/abs/2309.06180} {Efficient memory management
  for large language model serving with pagedattention}.

\bibitem[{Lefaudeux et~al.(2022)Lefaudeux, Massa, Liskovich, Xiong, Caggiano,
  Naren, Xu, Hu, Tintore, Zhang, Labatut, and Haziza}]{xFormers2022}
Benjamin Lefaudeux, Francisco Massa, Diana Liskovich, Wenhan Xiong, Vittorio
  Caggiano, Sean Naren, Min Xu, Jieru Hu, Marta Tintore, Susan Zhang, Patrick
  Labatut, and Daniel Haziza. 2022.
\newblock \href {https://github.com/facebookresearch/xformers} {xformers: A
  modular and hackable transformer modelling library}.

\bibitem[{Li et~al.(2023)Li, Zhao, Yu, Song, Li, Yu, Li, Huang, and
  Li}]{li2023apibank}
Minghao Li, Yingxiu Zhao, Bowen Yu, Feifan Song, Hangyu Li, Haiyang Yu, Zhoujun
  Li, Fei Huang, and Yongbin Li. 2023.
\newblock \href {http://arxiv.org/abs/2304.08244} {Api-bank: A comprehensive
  benchmark for tool-augmented llms}.

\bibitem[{Liu et~al.(2023)Liu, Lin, Hewitt, Paranjape, Bevilacqua, Petroni, and
  Liang}]{liu2023lost}
Nelson~F. Liu, Kevin Lin, John Hewitt, Ashwin Paranjape, Michele Bevilacqua,
  Fabio Petroni, and Percy Liang. 2023.
\newblock \href {http://arxiv.org/abs/2307.03172} {Lost in the middle: How
  language models use long contexts}.

\bibitem[{Lu et~al.(2022)Lu, Mishra, Xia, Qiu, Chang, Zhu, Tafjord, Clark, and
  Kalyan}]{lu2022learn}
Pan Lu, Swaroop Mishra, Tanglin Xia, Liang Qiu, Kai-Wei Chang, Song-Chun Zhu,
  Oyvind Tafjord, Peter Clark, and Ashwin Kalyan. 2022.
\newblock Learn to explain: Multimodal reasoning via thought chains for science
  question answering.
\newblock \emph{Advances in Neural Information Processing Systems},
  35:2507--2521.

\bibitem[{Lu et~al.(2023{\natexlab{a}})Lu, Peng, Cheng, Galley, Chang, Wu, Zhu,
  and Gao}]{lu2023chameleon}
Pan Lu, Baolin Peng, Hao Cheng, Michel Galley, Kai-Wei Chang, Ying~Nian Wu,
  Song-Chun Zhu, and Jianfeng Gao. 2023{\natexlab{a}}.
\newblock \href
  {https://proceedings.neurips.cc/paper_files/paper/2023/file/871ed095b734818cfba48db6aeb25a62-Paper-Conference.pdf}
  {Chameleon: Plug-and-play compositional reasoning with large language
  models}.
\newblock In \emph{Advances in Neural Information Processing Systems},
  volume~36, pages 43447--43478. Curran Associates, Inc.

\bibitem[{Lu et~al.(2023{\natexlab{b}})Lu, Qiu, Chang, Wu, Zhu, Rajpurohit,
  Clark, and Kalyan}]{lu2022dynamic}
Pan Lu, Liang Qiu, Kai-Wei Chang, Ying~Nian Wu, Song-Chun Zhu, Tanmay
  Rajpurohit, Peter Clark, and Ashwin Kalyan. 2023{\natexlab{b}}.
\newblock Dynamic prompt learning via policy gradient for semi-structured
  mathematical reasoning.
\newblock In \emph{International Conference on Learning Representations
  (ICLR)}.

\bibitem[{Milakov and Gimelshein(2018)}]{milakov2018online}
Maxim Milakov and Natalia Gimelshein. 2018.
\newblock Online normalizer calculation for softmax.
\newblock \emph{arXiv preprint arXiv:1805.02867}.

\bibitem[{oneDNN Contributors(2023)}]{oneapisr13:online}
oneDNN Contributors. 2023.
\newblock oneapi deep neural network library (onednn).
\newblock \url{https://github.com/oneapi-src/oneDNN}.

\bibitem[{OpenAI(2023{\natexlab{a}})}]{ChatGPTPlugins:online}
OpenAI. 2023{\natexlab{a}}.
\newblock Chatgpt plugins.
\newblock \url{https://platform.openai.com/docs/plugins/introduction}.

\bibitem[{OpenAI(2023{\natexlab{b}})}]{ChatGPTFuncCallDoc:online}
OpenAI. 2023{\natexlab{b}}.
\newblock Function calling - openai api.
\newblock \url{https://platform.openai.com/docs/guides/function-calling}.

\bibitem[{OpenAI(2023{\natexlab{c}})}]{gpt4}
OpenAI. 2023{\natexlab{c}}.
\newblock \href {https://doi.org/10.48550/arXiv.2303.08774} {Gpt-4 technical
  report}.
\newblock \emph{arXiv preprint arXiv:2303.08774}.

\bibitem[{OpenAI(2023{\natexlab{d}})}]{ChatGPTFuncCallBlog:online}
OpenAI. 2023{\natexlab{d}}.
\newblock How to call functions with chat models.
\newblock
  \url{https://cookbook.openai.com/examples/how_to_call_functions_with_chat_models}.

\bibitem[{OpenAI(2023{\natexlab{e}})}]{openaitiktok:online}
OpenAI. 2023{\natexlab{e}}.
\newblock openai/tiktoken: tiktoken is a fast bpe tokeniser for use with
  openai's models.
\newblock \url{https://github.com/openai/tiktoken}.

\bibitem[{Qian et~al.(2023)Qian, Han, Fung, Qin, Liu, and Ji}]{qian2023creator}
Cheng Qian, Chi Han, Yi~R. Fung, Yujia Qin, Zhiyuan Liu, and Heng Ji. 2023.
\newblock \href {http://arxiv.org/abs/2305.14318} {Creator: Tool creation for
  disentangling abstract and concrete reasoning of large language models}.

\bibitem[{Radford et~al.(2018)Radford, Narasimhan, Salimans, Sutskever
  et~al.}]{gpt1}
Alec Radford, Karthik Narasimhan, Tim Salimans, Ilya Sutskever, et~al. 2018.
\newblock Improving language understanding by generative pre-training.

\bibitem[{Radford et~al.(2019)Radford, Wu, Child, Luan, Amodei, Sutskever
  et~al.}]{gpt2}
Alec Radford, Jeffrey Wu, Rewon Child, David Luan, Dario Amodei, Ilya
  Sutskever, et~al. 2019.
\newblock Language models are unsupervised multitask learners.
\newblock \emph{OpenAI blog}, 1(8):9.

\bibitem[{Saad-Falcon et~al.(2023)Saad-Falcon, Barrow, Siu, Nenkova, Yoon,
  Rossi, and Dernoncourt}]{saadfalcon2023pdftriage}
Jon Saad-Falcon, Joe Barrow, Alexa Siu, Ani Nenkova, David~Seunghyun Yoon,
  Ryan~A. Rossi, and Franck Dernoncourt. 2023.
\newblock \href {http://arxiv.org/abs/2309.08872} {Pdftriage: Question
  answering over long, structured documents}.

\bibitem[{Schick et~al.(2023)Schick, Dwivedi-Yu, Dessì, Raileanu, Lomeli,
  Zettlemoyer, Cancedda, and Scialom}]{schick2023toolformer}
Timo Schick, Jane Dwivedi-Yu, Roberto Dessì, Roberta Raileanu, Maria Lomeli,
  Luke Zettlemoyer, Nicola Cancedda, and Thomas Scialom. 2023.
\newblock \href {http://arxiv.org/abs/2302.04761} {Toolformer: Language models
  can teach themselves to use tools}.

\bibitem[{Sheng et~al.(2023)Sheng, Zheng, Yuan, Li, Ryabinin, Fu, Xie, Chen,
  Barrett, Gonzalez et~al.}]{flexgen}
Ying Sheng, Lianmin Zheng, Binhang Yuan, Zhuohan Li, Max Ryabinin, Daniel~Y Fu,
  Zhiqiang Xie, Beidi Chen, Clark Barrett, Joseph~E Gonzalez, et~al. 2023.
\newblock High-throughput generative inference of large language models with a
  single gpu.
\newblock \emph{arXiv preprint arXiv:2303.06865}.

\bibitem[{Silfa et~al.(2022)Silfa, Arnau, and Gonz\'{a}lez}]{ebatch}
Franyell Silfa, Jose~Maria Arnau, and Antonio Gonz\'{a}lez. 2022.
\newblock \href {https://doi.org/ebatch} {E-batch: Energy-efficient and
  high-throughput rnn batching}.
\newblock \emph{ACM Trans. Archit. Code Optim.}, 19(1).

\bibitem[{Touvron et~al.(2023{\natexlab{a}})Touvron, Lavril, Izacard, Martinet,
  Lachaux, Lacroix, Rozi{\`e}re, Goyal, Hambro, Azhar
  et~al.}]{touvron2023llama}
Hugo Touvron, Thibaut Lavril, Gautier Izacard, Xavier Martinet, Marie-Anne
  Lachaux, Timoth{\'e}e Lacroix, Baptiste Rozi{\`e}re, Naman Goyal, Eric
  Hambro, Faisal Azhar, et~al. 2023{\natexlab{a}}.
\newblock Llama: Open and efficient foundation language models.
\newblock \emph{arXiv preprint arXiv:2302.13971}.

\bibitem[{Touvron et~al.(2023{\natexlab{b}})Touvron, Lavril, Izacard, Martinet,
  Lachaux, Lacroix, Rozière, Goyal, Hambro, Azhar, Rodriguez, Joulin, Grave,
  and Lample}]{llama}
Hugo Touvron, Thibaut Lavril, Gautier Izacard, Xavier Martinet, Marie-Anne
  Lachaux, Timothée Lacroix, Baptiste Rozière, Naman Goyal, Eric Hambro,
  Faisal Azhar, Aurelien Rodriguez, Armand Joulin, Edouard Grave, and Guillaume
  Lample. 2023{\natexlab{b}}.
\newblock \href {http://arxiv.org/abs/2302.13971} {Llama: Open and efficient
  foundation language models}.

\bibitem[{Trebino(2016)}]{poolallocator}
Mariano Trebino. 2016.
\newblock mtrebi/memory-allocators: Custom memory allocators in c++ to improve
  the performance of dynamic memory allocation.
\newblock \url{https://github.com/mtrebi/memory-allocators#pool-allocator}.

\bibitem[{Wei et~al.(2022)Wei, Wang, Schuurmans, Bosma, Xia, Chi, Le, Zhou
  et~al.}]{wei2022chain}
Jason Wei, Xuezhi Wang, Dale Schuurmans, Maarten Bosma, Fei Xia, Ed~Chi, Quoc~V
  Le, Denny Zhou, et~al. 2022.
\newblock Chain-of-thought prompting elicits reasoning in large language
  models.
\newblock \emph{Advances in Neural Information Processing Systems},
  35:24824--24837.

\bibitem[{White et~al.(2023)White, Fu, Hays, Sandborn, Olea, Gilbert, Elnashar,
  Spencer-Smith, and Schmidt}]{white2023prompt}
Jules White, Quchen Fu, Sam Hays, Michael Sandborn, Carlos Olea, Henry Gilbert,
  Ashraf Elnashar, Jesse Spencer-Smith, and Douglas~C. Schmidt. 2023.
\newblock \href {http://arxiv.org/abs/2302.11382} {A prompt pattern catalog to
  enhance prompt engineering with chatgpt}.

\bibitem[{Williams et~al.(2009)Williams, Waterman, and Patterson}]{roofline}
Samuel Williams, Andrew Waterman, and David Patterson. 2009.
\newblock \href {https://doi.org/10.1145/1498765.1498785} {Roofline: An
  insightful visual performance model for multicore architectures}.
\newblock \emph{Commun. ACM}, 52(4):65--76.

\bibitem[{Yu et~al.(2022)Yu, Jeong, Kim, Kim, and Chun}]{orca}
Gyeong-In Yu, Joo~Seong Jeong, Geon-Woo Kim, Soojeong Kim, and Byung-Gon Chun.
  2022.
\newblock Orca: A distributed serving system for $\{$Transformer-Based$\}$
  generative models.
\newblock In \emph{16th USENIX Symposium on Operating Systems Design and
  Implementation (OSDI 22)}, pages 521--538.

\bibitem[{Zhou et~al.(2023)Zhou, Muresanu, Han, Paster, Pitis, Chan, and
  Ba}]{zhou2023large}
Yongchao Zhou, Andrei~Ioan Muresanu, Ziwen Han, Keiran Paster, Silviu Pitis,
  Harris Chan, and Jimmy Ba. 2023.
\newblock \href {http://arxiv.org/abs/2211.01910} {Large language models are
  human-level prompt engineers}.

\bibitem[{Zhuang et~al.(2023)Zhuang, Yu, Wang, Sun, and
  Zhang}]{zhuang2023toolqa}
Yuchen Zhuang, Yue Yu, Kuan Wang, Haotian Sun, and Chao Zhang. 2023.
\newblock \href {http://arxiv.org/abs/2306.13304} {Toolqa: A dataset for llm
  question answering with external tools}.

\end{thebibliography}

\newpage

\appendix

\onecolumn
\section{System Prompt for Chatbot Applications with Plugins} \label{app:sys_prompt}

The following system prompt teaches the chatbot to use Bing Search, Expedia, OpenTable, and Spotify APIs to answer user queries. The token count is 1766.

\newtcolorbox{mybox}[2][]
  {colback = white, colframe = black, fonttitle = \fontfamily{ptm}\selectfont\bfseries\color{black},
    colbacktitle = white, enhanced, boxrule=0.5pt, titlerule=0.5pt,breakable,
    attach boxed title to top center={yshift=-2mm}, sharp corners,
    title=#2, boxed title style={sharp corners}, #1}

\begin{mybox}[colback=white]{System Prompt and User Query}
{\small Instructions: Given the following list of API specifications and user query, you will choose the most appropriate API to invoke and try to parse the corresponding parameters from the user query.

\begin{itemize}[leftmargin=*,noitemsep]
    \item If none of the API descriptions match the user query intent, you will return \texttt{not\_found()}.
    \item If a parameter is required but not included in the user query, then return  \texttt{not\_found()}.
    \item Your response must strictly follow the syntax of: \texttt{api\_chosen(param1=PARSED\_PARAM1, ...)}.
\end{itemize}

Following are the list of API definitions and their parameters:

\begin{itemize}[leftmargin=*]
    \item \texttt{bing\_web\_search(count, offset, q, safe\_search, set\_lang)}: The Web search API lets you send a search query to Bing and get back search results that include links to webpages, images, and more. If the user explicitly or implicitly wants to find the latest information from the web, you must use this API. \\
    Parameters:
    \begin{itemize}[leftmargin=*, label=-]
        \item \texttt{count}: [optional] The number of search results to return in the response. The default is 10 and the maximum value is 50.
        \item \texttt{offset}: [optional] The zero-based offset that indicates the number of search results to skip before returning results.
        \item \texttt{q}: [required] The user search query term. The term may not be empty.
        \item \texttt{safe\_search}: [optional] A filter used to filter results for adult content. ``Off'': Return webpages with adult text, images, or videos. ``Moderate'': Return webpages with adult text, but not adult images or videos. ``Strict'': Do not return webpages with adult text, images, or videos. The default is ``Moderate''.
        \item \texttt{set\_lang}: [optional] The language to use for user interface strings. You may specify the language using either a 2-letter or 4-letter code. Using 4-letter codes is preferred.
    \end{itemize}
    
    \item \texttt{bing\_images\_search(count, offset, q, safe\_search, set\_lang)}: The Image Search API lets you send a search query to Bing and get back a list of relevant images.\\
    Parameters:
    \begin{itemize}[leftmargin=*, label=-]
        \item \texttt{count}: [optional] The number of image results to return in the response. The actual number delivered may be less than requested. The default is 35. The maximum is 150.
        \item \texttt{offset}: [optional] The zero-based offset that indicates the number of image results to skip before returning results.
        \item \texttt{q}: [required] The user's search query term. The term may not be empty. The term may contain Bing Advanced Operators. For example, to limit images to a specific domain, use the ``site:'' operator.
        \item \texttt{safe\_search}: [optional] A filter used to filter results for adult content. ``Off'': Return webpages with adult text, images, or videos. ``Moderate'': Return webpages with adult text, but not adult images or videos. ``Strict'': Do not return webpages with adult text, images, or videos. The default is ``Moderate''.
        \item \texttt{set\_lang}: [optional] The language to use for user interface strings. You may specify the language using either a 2-letter or 4-letter code. Using 4-letter codes is preferred.
    \end{itemize}
    \item \texttt{expedia\_search\_hotel(city, hotel\_name, price\_buckets, star\_ratings, guest\_ratings)}: Search for a hotel based on the user query.\\
    Parameters:
    \begin{itemize}[leftmargin=*, label=-]
        \item \texttt{city}: [required] A string to identify the city to search for.
        \item \texttt{hotel\_name}: [optional] Hotel name used to filter the search results.
        \item \texttt{price\_buckets}: [optional] Used to define custom price filter buckets. If not provided then the default price filter buckets for the POS will be used.
        \item \texttt{star\_ratings}: [optional] Used to filter by star rating. Must be in increments of 5 and separated by commas (minStarRating=0 and maxStarRating=50). Ex. ``0,5,10'' means 0, 0.5 and 1 star hotels.
        \item \texttt{guest\_ratings}: [optional] Used to filter by amenities. Must be as a list of amenity ids separated by commas. Please note that there is no way at this time to validate the ids are actually valid.
    \end{itemize}
    \item \texttt{expedia\_search\_flights(departure\_date, return\_date, departure\_airport, arrival\_airport, number\_of\_adult\_travelers, child\_traveler\_age, non\_stop\_flight, airline\_preference)}: Flight search for one-way and round-trip scenarios.\\
    Parameters:
    \begin{itemize}[leftmargin=*, label=-]
        \item \texttt{departure\_date}: [required] Date the customer wants to leave for their flight on, in ISO format. 
        \item \texttt{return\_date}: [optional] Date the customer wants to return on. If present, indicates a round trip search. If not supplied, then it's a one-way search.
        \item \texttt{departure\_airport}: [required] The three-letter airport code for where the customer is leaving from.
        \item \texttt{arrival\_airport}: [optional] The three-letter airport code to where the customer is going. 
        \item \texttt{number\_of\_adult\_travelers}: [optional] Number of Adult Travelers (Default: 1).
        \item \texttt{child\_traveler\_age}: [optional] ``childTravelerAge'' represents the age of a single child traveler. You are required to specify the age of all child travelers. That means you must specify this parameter for each child that will be flying. Valid values are 0-17 (0 for an infant under the age of one year). If you would like to specify 3 child travelers with the ages of 1, 3 and 8 years, then your query string should contain: ``childTravelerAge=1\&childTravelerAge=3\&childTravelerAge=8''
        \item \texttt{non\_stop\_flight}: [optional] Set to true to return only nonstop flights in the search response (Default: False).
        \item \texttt{airline\_preference}: [optional] Optional parameter to get specific airline carrier information. By default, the preference is all.
    \end{itemize}
    \item  \texttt{opentable\_search\_restaurants(name, category, city, day)}: Returns a list of restaurants. \\
    Parameters:
    \begin{itemize}[leftmargin=*, label=-]
        \item \texttt{name}: [optional] Name of the restaurant to search for.
        \item \texttt{category}: [optional] Category of the restaurant to search for.
        \item \texttt{city}: [optional] City to search in.
        \item \texttt{day}: [optional] Date to search for.
    \end{itemize}
    \item  \texttt{spotify\_search\_catalog(q, type, limit, offset)}: Get Spotify Catalog information about albums, artists, playlists, tracks, shows or episodes that match a keyword string.\\
    Parameters:
    \begin{itemize}[leftmargin=*, label=-]
        \item \texttt{q}: [required] Search query. Keywords and optional field filters and operators.
        \item \texttt{type}: [required] A comma-separated list of item types to search across. Valid types are: album, artist, playlist, track, show and episode. Search results include hits from all the specified item types. For example: ``q=name:abacab\&type=album,track'' returns both albums and tracks with ``abacab'' included in their name.
        \item \texttt{limit}: [optional] Maximum number of results to return. Default: 20 Minimum: 1 Maximum: 50. Note: The limit is applied within each type, not on the total response. For example, if the limit value is 3 and the type is ``artist,album'', the response contains 3 artists and 3 albums.
        \item \texttt{offset}: [optional] The index of the first result to return. Default: 0 (the first result). Maximum offset (including limit): 1,000. Use with limit to get the next page of search results.
    \end{itemize}
\end{itemize}
  
Below are some examples of choosing the API that matches the user query:\\

datetime\_now=2023-11-17T10:45:07+08:00\\
user\_query=Do you believe in God?\\
api\_call=\texttt{not\_found()}\\

datetime\_now=2023-11-17T10:50:00+08:00\\
user\_query=What is the price of the iPhone 15 Pro Max?\\
api\_call=\texttt{bing\_web\_search(q=``price of iPhone 15 Pro Max'', set\_lang=``en'')} \\
  
datetime\_now=2023-11-17T11:09:10+08:00\\
user\_query=OpenAI's logo\\
api\_call=\texttt{bing\_images\_search(q=``OpenAI logo'', set\_lang=``zh'')}\\
  
datetime\_now=2023-11-17T13:21:30+08:00\\
user\_query=What is Taylor Swift's latest album?\\
api\_call=\texttt{spotify\_search\_catalog(q=``Taylor Swift'', type=``album'', limit=1)}\\
  
datetime\_now=2023-11-17T11:21:42+08:00\\
user\_query=Looking to eat vegan food in San Francisco this weekend, could you get me one great restaurant suggestion for Saturday?\\
api\_call=}
\end{mybox}

\end{document}